\documentclass[letterpaper, 10 pt, conference]{ieeeconf}  % Comment this line out if you need a4paper

\IEEEoverridecommandlockouts                              % This command is only needed if 
\usepackage{graphicx} % for including graphics
\usepackage{times} % assumes new font selection scheme installed
\usepackage{amsmath} % assumes amsmath package installed
\usepackage{amssymb}  % assumes amsmath package installed
\usepackage{threeparttable}
\usepackage{booktabs}
\usepackage{siunitx}
\usepackage{pifont} % 勾/叉
\newcommand{\cmark}{\ding{51}}
\newcommand{\xmark}{\ding{55}}

\title{\LARGE \bf
Anatomical Prior-Driven Framework for Autonomous Robotic Cardiac Ultrasound Standard View Acquisition}

\author{Zhiyan Cao$^{1,\dagger}$, Zhengxi Wu$^{2,\dagger}$, Yiwei Wang$^{1,4*}$, Pei-Hsuan Lin$^3$, Li Zhang$^5$, Zhen Xie$^6$, Huan Zhao$^{1}$, Han Ding$^{1}$
\thanks{*Corresponding author: Yiwei Wang}
\thanks{$^{1}$The authors are with the State Key Laboratory of Intelligent Manufacturing Equipment and Technology, Huazhong University of Science and Technology, Wuhan, Hubei 430074, China. 
        {\tt\small caozhiyan@hust.edu.cn, wang\_yiwei@hust.edu.cn, huanzhao@hust.edu.cn, dinghan@hust.edu.cn}}%
\thanks{$^{2}$Zhengxi Wu is with the School of Biomedical Engineering, Harbin Institute of Technology (Shenzhen), Shenzhen, Guangdong 518055, China. {\tt\small e1350337@u.nus.edu }}%
\thanks{$^{3}$Pei-Hsuan Lin is with the Information Intelligence Lab, Department of Electrical Engineering, National Chung Hsing University, Taichung 402, Taiwan. {\tt\small linl90050@gmail.com }}%
\thanks{$^{4}$Yiwei Wang is also with the Institute of Medical Equipment Science and Engineering, Huazhong University of Science and Technology, Luoyu Road 1037, Wuhan, Hubei 430074, China.}%
\thanks{$^{5}$Li Zhang is with the Department of Ultrasound Medicine, Union Hospital, Tongji Medical College, Huazhong University of Science and Technology, Jiefang Avenue 1277, Wuhan, Hubei 430022, China. {\tt\small 
zli429@hust.edu.cn }}%
\thanks{$^{6}$Zhen Xie is with the Institute of Systems Science (ISS),
National University of Singapore (NUS), Singapore 119615, Singapore. {\tt\small 
xav.xie@nus.edu.sg }}%
\thanks{$^\dagger$These authors contributed equally to this work and share the first authorship.}
\thanks{Accepted for publication at the IEEE International Conference on Robotics and Automation (ICRA), 2026.}
}

\begin{document}

\maketitle
\thispagestyle{empty}
\pagestyle{empty}

%%%%%%%%%%%%%%%%%%%%%%%%%%%%%%%%%%%%%%%%%%%%%%%%%%%%%%%%%%%%%%%%%%%%%%%%%%%%%%%%

\begin{abstract}

Cardiac ultrasound diagnosis is critical for cardiovascular disease assessment, but acquiring standard views remains highly operator-dependent. 
Existing medical segmentation models often yield anatomically inconsistent results in images with poor textural differentiation between distinct feature classes, while autonomous probe adjustment methods either rely on simplistic heuristic rules or black-box learning.
To address these issues, our study proposed an anatomical prior (AP)-driven framework integrating cardiac structure segmentation and autonomous probe adjustment for standard view acquisition.
A YOLO-based multi-class segmentation model augmented by a spatial-relation graph (SRG) module is designed to embed AP into the feature pyramid. 
Quantifiable anatomical features of standard views are extracted. Their priors are fitted to Gaussian distributions to construct probabilistic APs. 
The probe adjustment process of robotic ultrasound scanning is formalized as a reinforcement learning (RL) problem, with the RL state built from real-time anatomical features and the reward reflecting the AP matching. 
Experiments validate the efficacy of the framework. The SRG-YOLOv11s improves mAP\textsubscript{50} by $11.3\%$ and mIoU by $6.8\%$ on the \textit{Special Case} dataset, while the RL agent achieves a $92.5\%$ success rate in simulation and $86.7\%$ in phantom experiments.

\end{abstract}

%%%%%%%%%%%%%%%%%%%%%%%%%%%%%%%%%%%%%%%%%%%%%%%%%%%%%%%%%%%%%%%%%%%%%%%%%%%%%%%%
\section{Introduction}

\begin{figure*}[t]
 \centering
 \vspace{5pt}
 \includegraphics[width=0.99\linewidth]{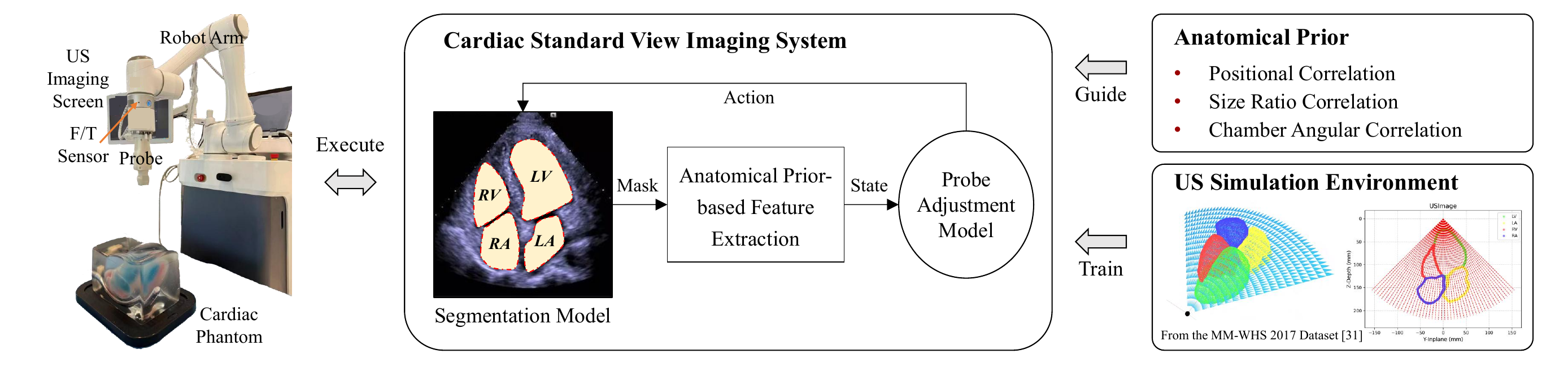}
 \vspace{-10pt}
 \caption{Framework of anatomical prior (AP)–driven robotic cardiac standard view acquisition system. 
 The AP standard takes the apical four-chamber (A4C) standard view as an example. 
 % For anatomical features between cardiac structures, the left atrium (LA) forms the posterior base of the heart and lies posterosuperior to the left ventricle (LV), whereas the right atrium (RA) and right ventricle (RV) occupy a more anterior aspect relative to the LA-LV complex \cite{mitchell2019guidelines}. For size ratios among chambers, LV end‐diastolic volume is typically on the order of about twice LA volume in healthy adults \cite{lang2015recommendations}.
 }
 \label{fig: framework}
\vspace{-15pt}
\end{figure*}

Cardiac ultrasound (US) diagnosis plays a pivotal role in the clinical assessment of cardiovascular diseases \cite{mitchell2019guidelines}. 
However, the acquisition of cardiac standard views, such as the apical four-chamber (A4C) view, which visualizes the right ventricle (RV), left ventricle (LV), right atrium (RA), and left atrium (LA) in a single plane, remains highly operator-dependent \cite{won2024sound}. 
It requires operators to simultaneously analyze the cardiac structure of the US image and adjust probe posture.
Variability in these two capabilities across operators leads to inconsistencies in view acquisition \cite{jiang2023robotic}.
To improve acquisition reproducibility, robotic US scanning has emerged as a promising paradigm \cite{bi2024machine}.
For such systems to match expert performance, they must achieve two tasks: accurate perception of cardiac structures and precise adjustment of probe posture.
A growing body of research has focused on these tasks \cite{tajbakhsh2020embracing, huang2024robot}, yet existing approaches still suffer from unreliable structure perception in US images \cite{li2024spatio} and unstable probe adjustment under clinical variability \cite{zakeri2024ai}. These limitations stem from insufficient utilization of anatomical priors (APs) \cite{painchaud2020cardiac, shida2023automated}, the inherent anatomical relationships of the heart (Fig. \ref{fig: framework}).

\begin{figure}[t]
 \centering
 \vspace{5pt}
 \includegraphics[width=0.95\linewidth]{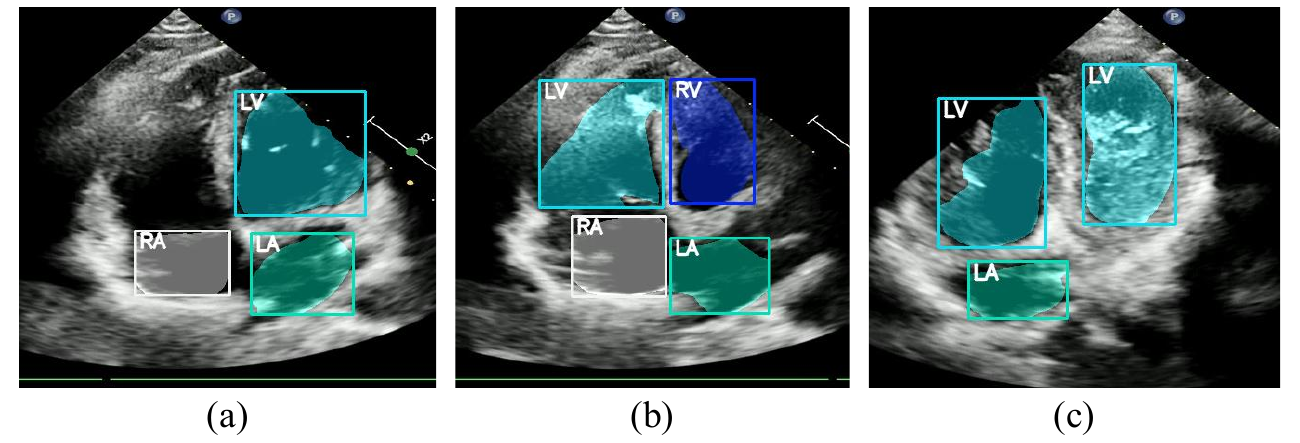}
 \vspace{-10pt}
 \caption{Typical segmentation challenges in cardiac US imaging. (a) represents missed detection of the RV when obscured by lung artifacts, (b) denotes mislabeling between the RV and LV, and (c) describes duplicate prediction of the LV.
 }
  \label{fig: segmentproblem}
\vspace{-15pt}
\end{figure}

Accurate segmentation of cardiac structures is the foundation of autonomous robotic US scanning \cite{jiang2023robotic}. 
Mainstream medical segmentation models are primarily appearance-based \cite{xie2021segformer}. They learn pixel-level patterns from image intensity and texture, but lack explicit integration of the anatomical constraints \cite{long2015fully}.
When faced with low-quality images, anatomical variability, or sparse annotations \cite{tajbakhsh2020embracing, petitjean2011review}, these models often produce anatomically inconsistent results \cite{painchaud2020cardiac, kong2021deep}.
Such limitations have motivated the incorporation of APs and topology-aware mechanisms.
% into the appearance-based segmentation pipelines. 
Some studies employed graph neural networks to encode contour-level topology \cite{li2024spatio}, some designed topological losses that penalize structural errors \cite{clough2020topological, berger2024topologically}, and others modeled inter-structure dependencies to leverage multi-target relations \cite{zhao2024farn, lu2025ap}.
However, in cardiac US structure segmentation scenarios, the images exhibit poor textural differentiation between distinct feature classes \cite{mazaheri2013echocardiography}.
Even with the integration of APs, existing methods still suffer from three issues in these scenarios (Fig. \ref{fig: segmentproblem}): missed detection of low-contrast or edge structures, mislabeling between texturally similar entities, and duplicate predictions caused by speckle noise.
% However, in the context of cardiac US structure segmentation scenarios, these methods still suffer from three critical issues: missed detection of low-contrast or edge structures (e.g., missing the RV when obscured by lung artifacts), mislabeling between classes of morphologically similar chambers (e.g., the LV being mistaken as the RV), and duplicate predictions caused by speckle noise (e.g., an LV cavity being detected as multiple instances).
% Even with AP-driven components, current 
% They struggle to accurately model and constrain the spatial relationships among cardiac structures.

Autonomous probe adjustment aims to align the probe from arbitrary initial postures to imaging cardiac standard views \cite{mitchell2019guidelines}.
Despite significant advancements in autonomous robotic US scanning \cite{jiang2023robotic}, their utilization of APs remains limited. 
Some studies explicitly incorporated APs by designing strategies based on anatomical feature feedback \cite{huang2024robot} or developing geometric-based algorithms \cite{van2025robotic}. However, these methods typically rely on simple heuristic rules, which are inadequate for complex, multi-faceted quality-evaluated tasks like acquiring cardiac standard views \cite{hao2025towards}.
Other studies designed learning-based methods by learning end-to-end correlations between US images and probe movements \cite{jiang2024sequence, jiang2024cardiac} or constructing the convolutional neural network (CNN) feature differences between real-time images and standard views to guide servo actuation \cite{zakeri2024ai, zakeri2025robust}.
However, the first type demands large-scale annotated data to ensure stability, where clinical high-quality labeled samples are scarce due to high labeling costs and privacy restrictions \cite{tajbakhsh2020embracing}.
The second type, leveraging feature differences for servo control, relies on CNN models trained exclusively on phantom data. This servo strategy limits their adaptation to the anatomical variability in clinical scenarios.

To address the above limitations, our study proposes an AP-driven cardiac standard view acquisition framework. This framework integrates cardiac structure segmentation and autonomous probe adjustment, with the A4C view as an example (Fig. \ref{fig: framework}).
The key insight of our framework is that APs can serve as a unifying bridge between the semantic segmentation of the US image and the probe adjustment for the robotic US scanning process. 
For the segmentation, APs impose explicit spatial–topological constraints that guide learning toward anatomically consistent predictions. For the probe adjustment, APs act as interpretable standard benchmarks to guide probe movements.
The framework follows a three-step workflow grounded in APs. First, it uses APs to enhance the reliability of cardiac structure segmentation. Second, quantifiable anatomical features were extracted from the segmentation results. Third, the autonomous probe adjustment is formulated as a reinforcement learning (RL) problem, where the RL state reflects real-time anatomical status and the reward function reflects the APs. The framework ensures robust structure perception and stable probe control, even under clinical variability of sim-to-real discrepancies.

The main contributions are listed as follows.

\begin{itemize}
 \item A spatial-relation graph (SRG)-augmented YOLO segmentation model is proposed, which embeds spatial-topological constraints, enhancing robustness against missed detection, mislabeling between classes, and duplicate prediction issues.
 \item An AP-guided RL problem for probe adjustment is developed, which overcomes the limitations of heuristic rules and black-box learning, facilitating simulation validation and zero-shot deployment on phantoms.
 \item An experimental platform for autonomous robotic acquisition of the cardiac standard A4C view is established, with its functionality validated through cardiac phantom experiments.
\end{itemize}

\section{Methods}

The proposed AP-driven framework consists of three sections.
In Sec. \ref{sec:seg}, the YOLO-based multi-class cardiac segmentation model, augmented with an SRG module, generates anatomically consistent masks from US images.
Then, quantifiable anatomical features are extracted from segmentation masks in Sec. \ref{sec: ap}. Priors of these features are fitted to Gaussian distributions to construct probabilistic APs, which serve as a quantifiable statistical benchmark for the anatomical features of standard views.
% encoding both the ideal anatomical state of standard views and acceptable physiological variability.
In Sec. \ref{sec: RL}, probe adjustment is formalized as a Markov decision process (MDP)-based RL problem. The RL state is constructed from current anatomical features, and its reward reflects the matching degree between these features and pre-fitted priors, ensuring interpretable, AP-aligned movements.

\begin{figure*}[ht]
 \centering
 \vspace{5pt}
 \includegraphics[width=0.99\linewidth]{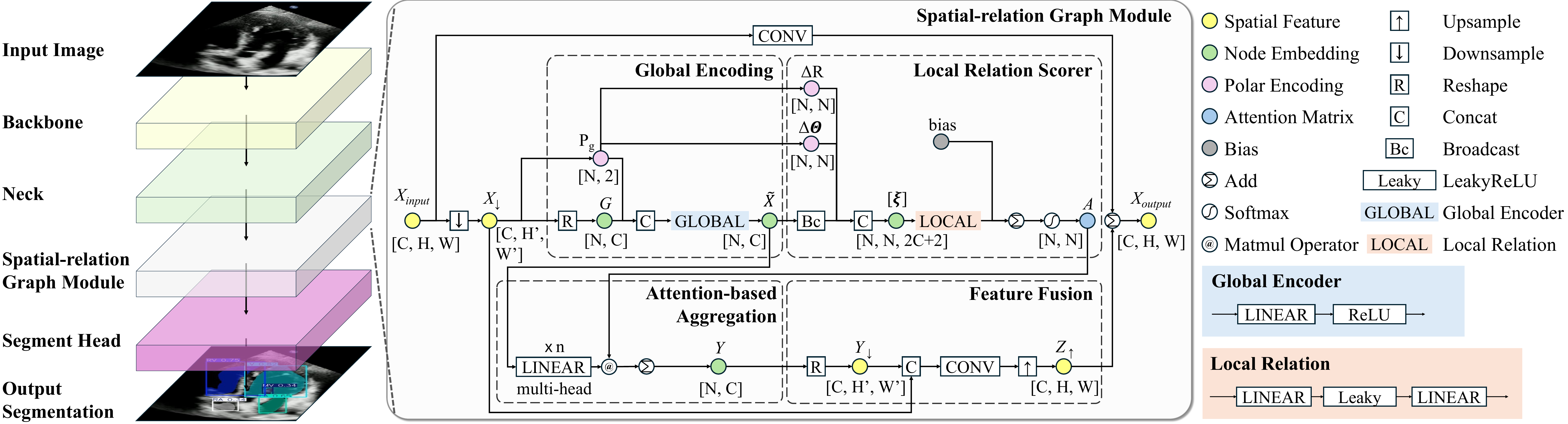}
 \vspace{-5pt}
 \caption{Framework of the YOLO-based multi-class cardiac segmentation model with the spatial-relation graph (SRG) module. Specifically, a backbone extracts semantic features from the input image. A multi-scale neck then fuses multi-resolution features to enhance the representation of cardiac structures. The SRG module subsequently embeds AP-guided constraints into the neck feature maps to promote anatomical consistency in the predictions. Finally, a segmentation head decodes the enhanced features into bounding boxes and instance segmentation masks.}
\label{fig: Graph-based Module}
\vspace{-15pt}
\end{figure*}

\subsection{Multi-class Cardiac Segmentation Model}
\label{sec:seg}

Reliable cardiac structure segmentation serves as the foundation for autonomous robotic US scanning \cite{jiang2023robotic}.
To enhance the reliability of cardiac structure segmentation, this section proposes a YOLO-based multi-class segmentation model augmented with the SRG module.

\subsubsection{Overall Structure}

The overall structure of the segmentation model includes a YOLO-style backbone, a multiscale neck, the SRG module, and a segment head, as illustrated in Fig. \ref{fig: Graph-based Module}.
Conventional YOLO models rely solely on image intensity and texture features, which could lead to anatomical errors (e.g., mislabeling LV as RV) in images with poor textural differentiation between distinct feature classes \cite{mazaheri2013echocardiography}. The SRG module is added to guide learning toward clinically plausible anatomical structures.

\subsubsection{Spatial-relation Graph Module}

The SRG module serves as the graph-based component for embedding AP-guided spatial-topological constraints into the segmentation model. Positioned between the multi-scale neck and segment head of the YOLO pyramid, it processes the neck-output features to model anatomical relationships between image regions.
This module is compatible with multiple YOLO variants, where the YOLOv11s is adopted as an instantiation in this section.

The workflow of the SRG module begins with adapting input features for graph-based modeling. The feature map $X_{\text{input}} \in \mathbb{R}^{C \times H \times W}$ (output by the neck) is first downsampled to $X_{\downarrow} \in \mathbb{R}^{C \times H' \times W'}$ (with $H' < H$ and $W' < W$). 
Here, $C$ denotes the number of feature channels, $H$ and $W$ represent the height and width of the input feature map, respectively.
The downsampled feature map $X_{\downarrow}$ is then reshaped into a graph structure $G \in \mathbb{R}^{N \times C}$, where each node $g_n$ represents a local image region and carries a $C$-dimensional feature vector encoding the appearance information of that region.

To inject global spatial context consistent with AP definitions, the SRG module performs global encoding on the graph $G$. For each node $g_n$ in the $(H', W')$ lattice, normalized polar coordinates $(r_{g_n}, \theta_{g_n})$ are computed to ensure global spatial information aligns with clinical descriptions of cardiac structure layout. These node-specific coordinates form a global spatial map $P_g = \{(r_{g_n}, \theta_{g_n}) \mid g_n \in \{g_1, \dots, g_N\}\}$, which is concatenated with the graph feature map $G$ to form a $(C+2)$-dimensional representation that fuses appearance and global spatial cues. 
A global encoder then processes this concatenated map into a globally encoded feature map $\widetilde{X} \in \mathbb{R}^{N \times C}$:
\begin{equation}
  \widetilde{X} = \mathrm{ReLU}\bigl([\; G \;\|\; P_g \;] \cdot W_g \bigr),
\end{equation}
where $\|$ denotes channel-wise concatenation and $W_g \in \mathbb{R}^{(C+2) \times C}$ is a linear layer.

Based on global encoding, the SRG module models fine-grained local anatomical relationships via a local relation scorer, which quantifies how anatomically relevant pairs of nodes are. For every pair of nodes $(p_n, q_n)$, two metrics are computed to capture AP-compliant local relationships: angular offset $\Delta\theta_{pq} = \operatorname{arctan2}\bigl(\sin(\theta_{p_n}-\theta_{q_n}),\,\cos(\theta_{p_n}-\theta_{q_n})\bigr)$ and radial offset $\Delta r_{pq} = r_{p_n}-r_{q_n}$.
These offsets directly correspond to the spatial constraints in APs: $\Delta\theta_{pq}$ describes the angular relationship between two regions (e.g., RV being anterior to LV) and $\Delta r_{pq}$ reflects their radial difference (e.g., LA being deeper than LV). 
To translate these offsets into actionable feature weights, the globally encoded matrix $\widetilde{X}$ is broadcast to a 3D tensor $\mathbb{R}^{N\times N\times C}$ to enable pairwise feature comparison, forming a pairwise descriptor $\xi_{pq}$ that combines the broadcasted features of $p_n$ and $q_n$ with $\Delta\theta_{pq}$ and $\Delta r_{pq}$. 
Specifically, write $\widetilde{X}=[\widetilde{x}_1,\dots,\widetilde{x}_N]^\top$ with each $\widetilde{x}_n\in\mathbb{R}^{C}$. Then define the broadcasted tensors $\widetilde{X}^{(p)}, \widetilde{X}^{(q)}\in\mathbb{R}^{N\times N\times C}$ by setting $\widetilde{X}^{(p)}_{pq} =\widetilde{x}_p$ and $\widetilde{X}^{(q)}_{pq} =\widetilde{x}_q$ for all pairs $(p,q)$.
A multi-layer perceptron with LeakyReLU activation function serves as the local relation function $\phi_e(\cdot)$, which is applied to $\xi_{pq}$ to output the raw affinity score $s_{pq}$:
\begin{equation}
 \xi_{pq} = \bigl[\widetilde{X}^{(p)}_{pq} \;\|\; \widetilde{X}^{(q)}_{pq} \;\|\; \Delta\theta_{pq} \;\|\; \Delta r_{pq}\bigr],
\end{equation} 
\begin{equation}
 s_{pq} = \phi_e(\xi_{pq}) = w_b^\top \cdot \mathrm{LeakyReLU}_{\alpha}\!\left(W_a \cdot \xi_{pq}\right),
\end{equation}
where $\mathrm{LeakyReLU}_{\alpha}(x)=\max(0,x)+\alpha\,\min(0,x)$, $\xi_{pq} \in \mathbb{R}^{2C+2}$, $W_a\!\in\!\mathbb{R}^{d\times(2C+2)}$ and $w_b\!\in\!\mathbb{R}^{d}$ are learnable parameters with a hidden width $d = C/2$.
A row-wise softmax operation on the adjusted scores $\hat{s}_{pq} = s_{pq} + b_{pq}$ generates an attention matrix $A \in \mathbb{R}^{N \times N}$, where each element $a_{pq} \in [0,1]$ represents the attention weight of node $q_n$ relative to $p_n$. This matrix acts as a dynamic adjacency graph that amplifies contributions from anatomically related nodes (e.g., anatomically adjacent nodes) and suppresses noise from irrelevant regions (e.g., background artifacts).

To integrate the above context, the SRG module employs an attention-based aggregation mechanism with a multi-head attention design. The attention matrix $A$ is shared across all heads to reduce parameter redundancy.
For each head $h$, a learnable projection matrix $W_h \in \mathbb{R}^{C \times C}$ transforms the globally encoded matrix $\widetilde{X}$ into head-specific feature values $V^{(h)} \in \mathbb{R}^{N \times C}$. Feature aggregation is then performed via matrix multiplication of $A$ and $V^{(h)}$, followed by ReLU activation to introduce non-linearity, yielding head-specific aggregated features $M^{(h)} \in \mathbb{R}^{N \times C}$. Features from all heads are summed to fuse multi-view anatomical context into a single aggregated matrix $M \in \mathbb{R}^{N \times C}$, and a final projection matrix $W_o \in \mathbb{R}^{C \times C}$ converts $M$ into $Y \in \mathbb{R}^{N \times C}$.

The final step of the SRG module is the feature fusion, which preserves fine-grained appearance details from the original input while retaining AP-constrained graph-enhanced features. The aggregated matrix $Y$ is reshaped back into a 2D feature map $Y_{\downarrow} \in \mathbb{R}^{C \times H' \times W'}$, matching the spatial dimensions of $X_{\downarrow}$ to enable channel-wise concatenation. The concatenated map $[X_{\downarrow} \parallel Y_{\downarrow}]$ is processed by a $1\times1$ convolution to re-weight channels, fusing appearance features (from $X_{\downarrow}$) and graph-enhanced anatomical features (from $Y_{\downarrow}$) at each spatial location while preserving spatial resolution.

The fused map $Z_{\downarrow} \in \mathbb{R}^{C \times H' \times W'}$ is then upsampled to the original resolution of $X_{\text{input}}$ and combined with a residual projection of $X_{\text{input}}$ (via a $1\times1$ convolution) to form the final output $X_{\text{output}} \in \mathbb{R}^{C \times H \times W}$. This residual connection preserves low-level structural features that are critical for the segmentation of small anatomical structures, while the upsampled fused features ensure anatomical constraints are maintained across the entire feature map. The $X_{\text{output}}$ is then passed to the segment head, where it is decoded into bounding boxes and instance segmentation masks. 
These masks provide reliable information for the anatomical feature extraction and autonomous probe adjustment.

\subsection{Anatomical Feature Extraction for Cardiac Standard View}
\label{sec: ap}

To translate the anatomical laws of the cardiac standard view into quantifiable statistical benchmarks, this section focuses on the extraction of anatomical features and the construction of their statistical priors.

To enable the extraction of anatomical features aligning with clinical standards, the cardiac entity sets are defined to organize the features in a standardized manner.
Let $\mathcal{C}$ denotes the total set of cardiac entities.
For a specific standard view, the view-included entity set ${\boldsymbol{c}_{\text{in}}} \subset \mathcal{C}$ represents entities that should be visualized in the standard view, and the view-excluded entity set $\boldsymbol{c}_{\text{ex}} \subset \mathcal{C} \setminus {\boldsymbol{c}_{\text{in}}}$ denotes entities that should not be visible in this view. 
For example, the A4C standard view has the view-included entity set $\boldsymbol{c}_{\text{in}} = \{\text{LV}, \text{RV}, \text{LA}, \text{RA}\}$ and view-excluded entity set $\boldsymbol{c}_{\text{ex}} = \{\text{aorta}\}$, according to the American Society of Echocardiography (ASE) guidelines \cite{mitchell2019guidelines}.

The anatomical features are extracted from the instance masks generated by the segmentation model, focusing on the relative position features and the size ratio features.
The relative position features capture the angular and radial constraints between pairs of target entities within the cardiac standard view.
The target entity pair set $\mathcal{P}_{\text{tar}} = \{(c_i, c_j) \mid c_i, c_j \in \boldsymbol{c}_{\text{in}}, c_i \neq c_j \}\in \mathbb{R}^{2\times m}$ is defined according to the anatomical connectivity between the two entities in each pair. For example, the target entity pair set of the A4C standard view is $\mathcal{P}_{\text{tar}} = \left\{ (\text{LV, RV}), (\text{LA, RA}), (\text{LV, LA}), (\text{RV, RA}) \right\}$, yielding $m=4$ pairs.
For each pair $(c_i, c_j) \in \mathcal{P}_{\text{tar}}$, the polar angle difference $\Delta \theta_{i,j} = \theta_{c_i} - \theta_{c_j}$ describes the angular offset between $c_i$ and $c_j$, where $\theta_{c}$ is the average polar angle of all pixels in the entity $c$. Meanwhile, the radial difference $\Delta r_{i,j} = r_{c_i} - r_{c_j}$ quantifies the radial offset, where $r_{c}$ is the average normalized radius of all pixels in the entity $c$. These values are organized into the angular offset vector ${\Delta\Theta}_{\text{in}} = [ \Delta\theta_{i,j} \mid (c_i, c_j) \in \mathcal{P}_{\text{tar}} ] ^\top$ and the radial offset vector ${\Delta R}_{\text{in}} = [ \Delta r_{i,j} \mid (c_i, c_j) \in \mathcal{P}_{\text{tar}} ] ^\top$.
The size ratio features reflect the stable volume proportion constraints between entities within the cardiac standard view.
Using the clinically recognized LV as the reference \cite{mitchell2019guidelines}, the pixel area $S_c$ of each entity $c \in \boldsymbol{c}_{\text{in}}$ is normalized by the area of LV $S_{\text{LV}}$ to a relative area ratio $\alpha_c = \frac{S_c}{S_{\text{LV}}}$. The ratios are organized into a size ratio vector $\boldsymbol{\alpha}_{\text{in}} = [\alpha_c \mid c \in \boldsymbol{c}_{\text{in}} ] \top$.

To convert real-time anatomical features into a quantifiable benchmark for probe adjustment, the priors of these features are constructed as the Gaussian distribution parameters that characterize the statistical properties in qualified standard views.
For the relative position features, each angular offset $\Delta\theta_{i,j}$ in $\Delta\Theta_{\text{in}}$ follows $\mathcal{N}(\mu_{i,j}^\theta, (\sigma_{i,j}^{\theta})^{2})$. Here, $\mu_{i,j}^\theta$ denotes the average angular offset of entity pair $(c_i, c_j)\in \mathcal{P}_{\text{tar}}$ in standard views, and $(\sigma_{i,j}^{\theta})^{2}$ denotes the squared standard deviation quantifying physiological variability.
Similarly, each radial offset $\Delta r_{i,j}$ in $\Delta R_{\text{in}}$ follows $\mathcal{N}(\mu_{i,j}^r, (\sigma_{i,j}^{r})^{2})$, with $\mu_{i,j}^r$ as the average radial offset and $(\sigma_{i,j}^{r})^{2}$ as its squared standard deviation.
For size ratio features, each $\alpha_c$ in $\boldsymbol{\alpha}_{\text{in}}$ follows $\mathcal{N}(\mu_c^\alpha, (\sigma_{c}^{\alpha})^{2})$, where $\mu_c^\alpha$ is the average area ratio of entity $c$ relative to LV in standard views, and $(\sigma_{c}^{\alpha})^{2}$ is its squared standard deviation.  
These Gaussian parameters are estimated from a prepared dataset. For any feature $f$ (e.g., $\Delta\theta_{i,j}$ or $\alpha_c$), given its sample set $\mathcal{D}_f = \{f_1, \dots, f_M\}$ from $M$ standard views, the mean (e.g., $\mu_{i,j}^\theta$, $\mu_c^\alpha$) is the arithmetic average of $\mathcal{D}_f$, and the squared standard deviation (e.g., $(\sigma_{i,j}^{\theta})^{2}$, $(\sigma_{c}^{\alpha})^{2}$) is the mean squared deviation from this mean.
Together, these parameters form an interpretable standard benchmark for guiding probe adjustment.

\subsection{Autonomous Probe Adjustment}

\label{sec: RL}

Acquiring the cardiac standard view requires precise US probe adjustment. 
Formulating autonomous probe adjustment as an RL problem under an MDP enables the agent to learn optimal probe strategies by integrating AP knowledge and interactive feedback. 
The MDP formulation is defined as $ \mathcal{M} = (\mathcal{S}, \mathcal{A}, \mathcal{T}, \mathcal{R})$, where $ \mathcal{S} $ is the state space, $ \mathcal{A} $ is the action space, $ \mathcal{T}: \mathcal{S} \times \mathcal{A} \times \mathcal{S} \rightarrow [0,1]$ is the transition function, $ \mathcal{R}: \mathcal{S} \times \mathcal{A} \rightarrow \mathbb{R} $ is the reward function. The goal is to learn a policy $ \pi: \mathcal{S} \rightarrow \mathcal{A} $ that maximizes the cumulative reward, guiding the US probe to the optimal standard view.

\subsubsection{State Definition}

When the probe is adjusted to a new posture, it captures a new US image $\textbf{I}(t)$ at the time step $t$, which is processed by the segmentation model to generate instance masks. 
These masks are fed into the AP feature extraction to compute the state $\textbf{s}(t) \in \mathcal{S}$. 
The state $\textbf{s}(t))$ integrates four metrics reflecting AP conformity: 
\begin{equation}
\textbf{s}(t) = [
\boldsymbol{\alpha}_{\text{in}}(t)^\top ,
\boldsymbol{\alpha}_{\text{ex}}(t)^\top ,
\phi_{\text{all}}(t) ,
{s}_{\text{position}}(t)  ]^\top, 
\end{equation}
where $\boldsymbol{\alpha}_{\text{in}}$ is the area ratio vector of the current image defined in Sec. \ref{sec: ap}, and $\boldsymbol{\alpha}_{\text{ex}} = [\alpha_c \mid c \in \boldsymbol{c}_{\text{ex}} ] \top$ represents the relative area of excluded entities $\boldsymbol{c}_{\text{ex}}$.
The global polar angle $\phi_{\text{all}}$ measures the deviation of the cardiac centroid from the ideal image center.

The position correlation feature $s_{\text{position}} \in [0,1]$ quantifies the consistency of position relationships in the current view with the prior:
\begin{equation}
\begin{aligned}
s_{\text{position}} &= \frac{1}{2\sum_{(c_i,c_j) \in \mathcal{P}_{\text{tar}}} w_{i,j} } \cdot
\\&\sum_{(c_i,c_j) \in \mathcal{P}_{\text{tar}}} w_{i,j} [p(\Delta\theta_{i,j} \mid \mu_{i,j}^\theta, (\sigma_{i,j}^{\theta})^{2}) 
\\& + p(\Delta r_{i,j} \mid \mu_{i,j}^r, (\sigma_{i,j}^{r})^{2}) ], 
\end{aligned}
\end{equation}
where $w_{i,j} \geq 0 $ are hyperparameters reflecting the clinical importance of each pair’s position relationship, $\Delta\theta_{i,j}$ and $\Delta r_{i,j}$ represent the measured angular and radial differences between chambers $c_i$ and $c_j$ in the current view, $\mu^\theta_{i,j}, (\sigma_{i,j}^{\theta})^{2}$ and $\mu^r_{i,j}, (\sigma_{i,j}^{r})^{2}$ are the prior mean and variance of these features, and $p(\cdot)$ denotes the probability density function of the Gaussian distribution.

\subsubsection{Action Space}

To focus on fine-grained control relevant to standard view acquisition, the adjustment of the US probe is constrained to a single degree of freedom. The proposed action space is discrete with seven fine-tuning operations $\mathcal{A} = \{a_0, a_1, a_2, a_3, a_4, a_5, a_6\}$. Here, actions $a_0$ and $a_1$ correspond to the US probe rotating along the $x$-axis with a positive or negative angle $\delta$, actions $a_2$ and $a_3$ represent rotation along the $y$-axis with a positive or negative angle $\delta$, actions $a_4$ and $a_5$ denote rotation along the $z$-axis with a positive or negative angle $\delta$, and action $a_6$ means maintaining the current posture, respectively.

At each time step $t$, the agent selects an action $a(t)$ from the action space $\mathcal{A}$, which is then executed by the system to interact with the cardiac model.

\subsubsection{Reward Function}

When a new state $\textbf{s}(t)$ is reached at the time step $t$, the RL agent receives a reward $r(t)$. The reward $r(t)$ is designed to evaluate the conformity of the current state to APs, integrating area ratio consistency of view-included entities $r_{\text{in}}$ and excluded entities $r_{\text{ex}}$, global polar angle rationality $|\phi_{\text{all}}|$, and position consistency $s_{\text{position}}$:
\begin{equation}
r(t) = w_1 \cdot r_{\text{in}} + w_2 \cdot r_{\text{ex}} + w_3 \cdot \vert \phi_{\text{all}} \vert + w_4 \cdot s_{\text{position}}, 
\label{Eq. reward}
\end{equation}
where $w_1, w_2, w_3, w_4 \geq 0$ are hyperparameters representing the importance of each metric.

The area ratio consistency metrics $r_{\text{in}}$ and $r_{\text{ex}}$ collectively quantify the alignment of the relative area ratios of cardiac entities with the prior.
For either entity set $\boldsymbol{c}\in \{\boldsymbol{c}_{\text{in}}, \boldsymbol{c}_{\text{ex}}\}$, the consistency of the area ratio is calculated as:
\begin{equation}
r_{\boldsymbol{c}} = \frac{1}{\sum_{c \in \boldsymbol{c}} w_c} \sum_{c \in \boldsymbol{c}} w_c p(\alpha_c),
\end{equation}
where $w_c \geq 0$ is the weight assigned to the entity $c$, and $p(\alpha_c)$ is the prior-aligned probability term adapted to the type of $c$:
\begin{equation}
p(\alpha_c) = 
\begin{cases} 
\frac{\mathcal{N}(\alpha_c \mid \mu^\alpha_{c}, (\sigma_{c}^{\alpha})^{2})}{\max\left( \mathcal{N}(\cdot \mid \mu^\alpha_{c}, (\sigma_{c}^{\alpha})^{2}) \right)} & \text{if } c \in \boldsymbol{c}_\text{in} \\
-\alpha_c & \text{if } c \in \boldsymbol{c}_\text{ex}.
\end{cases}
\end{equation}

The RL agent is first trained in a cardiac simulation environment (Fig. \ref{fig: framework}) based on the proposed MDP formulation, and then deployed to the real-world experiment scenario for practical validation.

\section{Experiments and Results}

To validate the proposed framework, experiments focus on evaluating the segmentation model’s performance in clinical cases (Sec. \ref{sec: exp_seg}) and validating the RL approach in robotic standard cardiac view acquisition deployment (Sec. \ref{sec: exp_RL}).
Each subsection details the corresponding setup and results.

\begin{figure*}[t]
 \centering
 \vspace{5pt}
 \includegraphics[width=0.95\linewidth]{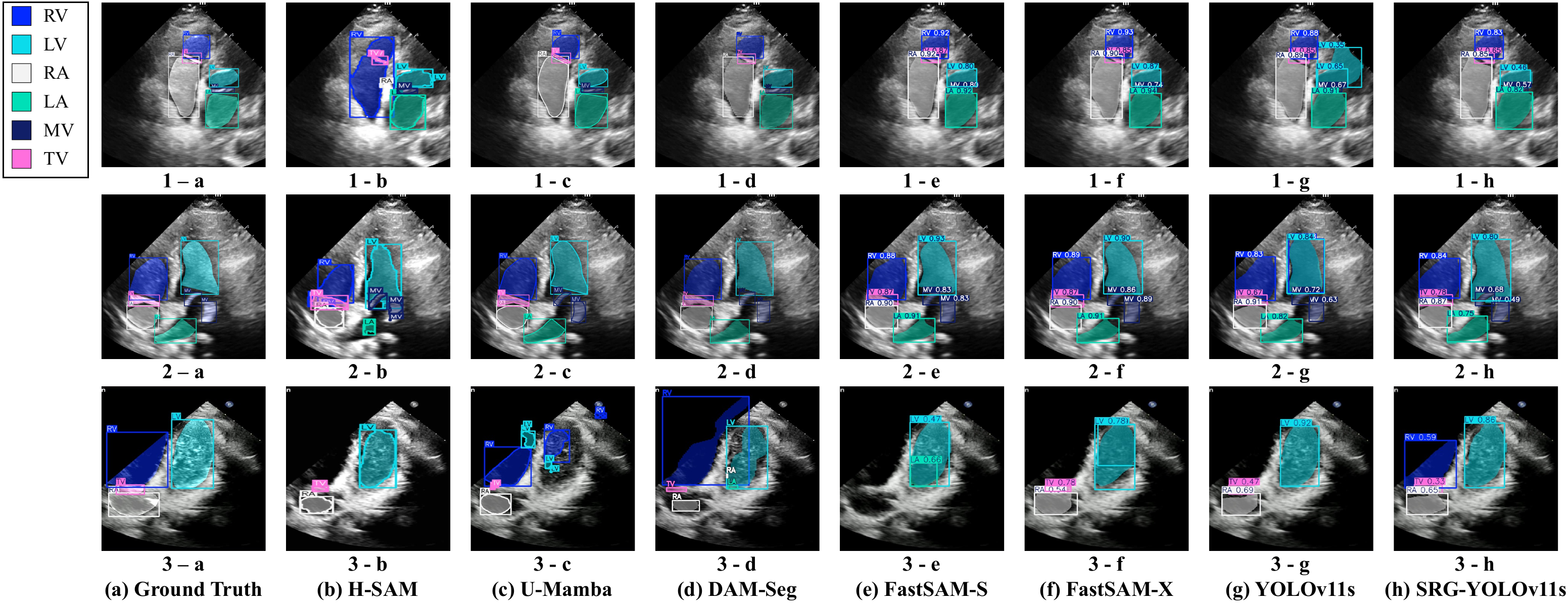}
 \vspace{-10pt}
 \caption{Visualized examples present scenarios involving the three critical errors in cardiac US segmentation: missed detection, mislabeling between classes, and duplicate predictions. Each row corresponds to a distinct segmentation scenario. The first column shows the ground truth, and each subsequent column displays the output of one evaluated model.
 As a result, the segmentation results in 3-b, 3-e, 3-f, and 3-g exhibit missed detection, the results in 1-b, 2-b, 2-g, 3-c, 3-d, and 3-e show mislabeling between classes, and the results in 1-b, 1-g, 2-b, 2-g, 3-c, 3-d, and 3-f have duplicate predictions. 
 Notably, the SRG-YOLOv11s does not exhibit any of these three critical errors across the scenarios.}
 \label{fig: experiment_1}
\vspace{-5pt}
\end{figure*}

\label{sec: exp}

\subsection{Segmentation Model Evaluation}

\label{sec: exp_seg}

\subsubsection{Evaluation Setup}

To validate the ability of the SRG module to address the challenges in US cardiac segmentation (missed detection, mislabeling between classes, duplicate prediction), a \textit{Special Case} evaluation dataset was curated by manually sampling 145 challenging A4C images from real-world US scanning videos acquired by physicians. 
The private dataset for the segmentation model training comprises 465 US images (415 for training, 50 for validation). Each image is annotated with 6 anatomical labels, namely the RV, LV, RA, LA, mitral valve (MV), and tricuspid valve (TV).
% Segmentation training and validation were performed in Python on an Intel Core i9-14900K workstation with an NVIDIA RTX 4090 GPU.

\subsubsection{Evaluation Results}

The proposed segmentation model was evaluated on the \textit{Special Case} dataset with four metrics: mAP\textsubscript{50}, mAP\textsubscript{50--95}, mIoU, and mDice. All the models were trained under a unified protocol, which included identical train and validation splits, fixed seeds, learning-rate schedule, and data augmentation strategies.

\begin{table}[t]
  \centering
  \vspace{5pt}
  \caption{Comparison experiments on the \textit{Special Case} dataset.}
  \label{tab:comparemodel}
  \vspace{-5pt}
  % \footnotesize
  \begin{threeparttable}
    \setlength{\tabcolsep}{7pt}
    \renewcommand{\arraystretch}{1}
    \begin{tabular}{@{}l
        S[table-format=1.3]
        S[table-format=1.3]
        S[table-format=1.3]
        S[table-format=1.3]@{}}
      \toprule
      Models        & {mAP\textsubscript{50}} & {mAP\textsubscript{50--95}} & {mIoU} & {mDice} \\
      \midrule
      YOLOv5s       & 0.397 & 0.146 & 0.411 & 0.490 \\
      YOLOv8s       & 0.352 & 0.128 & 0.397 & 0.478 \\
      YOLOv11s      & 0.423 & 0.169 & 0.426 & 0.506 \\
      FastSAM-S     & 0.418 & 0.144 & 0.373 & 0.456 \\
      FastSAM-X     & 0.407 & 0.165 & 0.357 & 0.427 \\
      H-SAM         & \multicolumn{1}{c}{--} & \multicolumn{1}{c}{--} & 0.261 & 0.334 \\
      U-Mamba (Bot) & \multicolumn{1}{c}{--} & \multicolumn{1}{c}{--} & 0.234 & 0.299 \\
      DAM-Seg       & \multicolumn{1}{c}{--} & \multicolumn{1}{c}{--} & 0.345 & 0.494 \\
      Ours (v5)     & 0.416 & 0.157 & 0.423 & 0.503 \\
      Ours (v8)     & 0.441 & 0.170 & 0.416 & 0.495 \\
      Ours (v11)    & {\bfseries 0.471} & {\bfseries 0.194} & {\bfseries 0.455} & {\bfseries 0.530} \\
      \bottomrule
    \end{tabular}
  \end{threeparttable}
\vspace{-15pt}
\end{table}

\begin{table*}[t]
  \centering
  \begin{threeparttable}
  \caption{Ablation on SRG-YOLOv11s: effect of the global encoding and local relation scorer.}
  \label{tab:ablation}
  \setlength{\tabcolsep}{13pt}
  \renewcommand{\arraystretch}{1}

  {\sisetup{detect-weight=true,detect-family=true}

  \begin{tabular}{@{}l c c
    S[table-format=1.3,table-number-alignment=center]  % Precision
    S[table-format=1.3,table-number-alignment=center]  % Recall
    S[table-format=1.3,table-number-alignment=center]  % mAP50
    S[table-format=1.3,table-number-alignment=center]  % mAP50--95
    S[table-format=1.3,table-number-alignment=center]  % mIoU
    S[table-format=1.3,table-number-alignment=center]  % mDice
  @{}}
    \toprule
    Variant        & Global & Local
                   & {Precision} & {Recall} & {mAP\textsubscript{50}} & {mAP\textsubscript{50--95}} & {mIoU} & {mDice} \\
    \midrule
    SRG-YOLOv11s    & \cmark & Full     & 0.471 & 0.480 & {\bfseries 0.471} & {\bfseries 0.194} & {\bfseries 0.455} & {\bfseries 0.530} \\
    SRG-YOLOv11s    & \xmark & Full     & 0.461 & {\bfseries 0.496} & 0.468 & 0.185 & 0.444 & 0.523 \\
    SRG-YOLOv11s    & \cmark & Identity & 0.392 & 0.368 & 0.359 & 0.141 & 0.407 & 0.485 \\
    SRG-YOLOv11s    & \xmark & Identity & 0.402 & 0.412 & 0.362 & 0.126 & 0.409 & 0.490 \\
    \midrule
    YOLOv11s-only    & None & None     & {\bfseries 0.485} & 0.433 & 0.423 & 0.169 & 0.426 & 0.506 \\
    \bottomrule
  \end{tabular}
  }% end local \sisetup

  \begin{tablenotes}[flushleft]
    \footnotesize
    \item \textbf{Global:} \cmark\ = global encoding enabled, \xmark\ = disabled. \textbf{Local:} \emph{Full} = local affinity modeling over the neighborhood graph,
          \emph{Identity} = identity mapping (self-connection only, no affinities).
          \textbf{None} = no SRG module.
  \end{tablenotes}
  \end{threeparttable}
\vspace{-15pt}
\end{table*}

Table~\ref{tab:comparemodel} summarizes the general results for all four evaluation metrics on the \textit{Special Case} dataset, covering comparisons among the SRG-augmented YOLO variants, their baselines (YOLOv5s/8s/11s \cite{ultralytics_models}), FastSAM\allowbreak-S/X \cite{zhao2023fast}, H-SAM \cite{cheng2024unleashing}, U-Mamba \cite{U-Mamba}, and DAM-Seg \cite{ullah2025dam}.
While a unified protocol ensured consistency across models, method-specific setups were still preserved.
H-SAM was implemented with its official release, which sets the input size to $512$. U-Mamba was trained via nnU-Net, which automatically configured the preprocessing and training patch sizes ($382\times640$ in our evaluation). For DPT-based DAM-Seg, the input resolution was fixed to $224\times224$ to preserve the ViT-Base ($16\times16$ patch size) token grid and its learned absolute positional embeddings. For H-SAM, U-Mamba, and DAM-Seg, the optimizer type and initial learning rate were retained as specified in the authors’ official releases.
Absolute scores were reported based on these method-specific constraints.
The evaluation results shown in Tab.~\ref{tab:comparemodel} validate that integrating the SRG module into YOLOv5s/8s/11s improves performance across all four metrics for each backbone, with the SRG-YOLOv11s notably leading across all four evaluation metrics.
Gains remained consistent under the stricter mAP\textsubscript{50--95}.
Compared to YOLOv11s, the SRG-augmented YOLOv11s improved mAP\textsubscript{50} by $11.3\%$, mAP\textsubscript{50--95} by $14.8\%$, mIoU by $6.8\%$, and mDice by $4.7\%$. Such improvements were also observed in the SRG-augmented YOLOv5s and YOLOv8s. Improvements under mAP\textsubscript{50--95} suggested better localization quality across a range of IoU thresholds, while improvements in mDice reflect sharper boundaries.

Besides, an ablation study was conducted under the same protocol and fixed seeds on the YOLOv11s backbone to isolately test the contributions of the global encoding and the local relation scorer.
As shown in Tab.~\ref{tab:ablation}, enabling both the SRG module’s global encoding and full local relation scorer yielded the best results on four metrics.
With only the SRG's global encoding ablated, the variant’s performance remained above the YOLOv11s-only baseline.
Replacing the full local relation scorer with an identity mapping degraded all metrics in both global encoding settings, underscoring the necessity of learned local relations.
Furthermore, while the local relation scorer alone already outperformed the baseline, enabling global encoding provided additional stable gains.
These results indicate that learned local relations and global encoding are complementary. The former yields clear improvements over the baseline, while the latter adds consistent gains, with a more apparent effect under stricter IoU thresholds.
At a fixed operating threshold, SRG variants trade a small decrease in precision for a larger increase in recall, which is a typical precision–recall trade-off. Nevertheless, the summary metrics improve, indicating an overall gain in detection and segmentation quality.

% Fig. \ref{fig: experiment_1} presents qualitative visualizations of A4C view segmentation, focusing on the three critical challenges (missed detection, mislabeling between classes, duplicate prediction). In the figure, baseline methods (e.g., YOLOv11s, FastSAM) consistently exhibit these errors across scenarios, while the SRG-YOLOv11s tackles these issues to maintain anatomical consistency.

Fig. \ref{fig: experiment_1} presents visualized comparisons of A4C view segmentation results across different models, focusing on the missed detection, mislabeling between classes, and duplicate prediction challenges. In the figure, baseline methods consistently exhibit these errors across scenarios, while only the SRG-YOLOv11s maintains anatomical consistency.

% Fig. \ref{fig: experiment_1} presents visualized comparisons on A4C frames, specifically showing qualitative examples of the three critical issues in cardiac US segmentation.
% Only the SRG-YOLOv11s does not exhibit any of these errors across all scenarios, ensuring anatomical consistency.

% These qualitative results align with the quantitative improvements of the SRG-YOLOv11s, further validating the effectiveness of AP-driven constraints to enhance segmentation reliability.

% Fig. \ref{fig: experiment_1} presents qualitative visualizations of A4C view segmentation, focusing on the three critical challenges (missed detection, cross-class mislabeling, duplicate prediction) that are prevalent in cardiac US segmentation. As shown in the figure, baseline methods (e.g., YOLOv11s, FastSAM) consistently exhibit these errors across scenarios, while the SRG-YOLOv11s tackles these issues effectively to maintain anatomical consistency. These qualitative results align with the quantitative improvements of the SRG-YOLOv11s, further validating the effectiveness of AP-driven constraints in enhancing segmentation reliability.

\begin{table}[t]
  \centering
  % \vspace{5pt} 
  \caption{Parameters of the MDP Simulation for RL Training.}
  \label{Tab: para}
  \vspace{-5pt}  % 与参考格式一致的caption下方间距
  \begin{threeparttable}
    \setlength{\tabcolsep}{6pt}  % 调整列间距（根据内容微调，比参考8pt略宽以适配6列）
    \renewcommand{\arraystretch}{1}  % 与参考格式一致的行高
    \begin{tabular}{@{}c c c c c c@{}}  % 6列：左对齐参数名+居中值，交替排列，消除首尾多余间距
      \toprule  % booktabs顶部粗线，替代原hline
      Param. & Value & Param. & Value & Param. & Value \\
      \midrule  % booktabs中间细线，分隔表头与内容
      $\delta$ & $1^\circ$ & $w_1$ & 0.7 & $w_2$ & 0.14 \\
      $w_3$ & 3 & $w_4$ & 0.1 & $\forall (i,j) \in \mathcal{P}_{\text{tar}}: w_{i,j}$ & 1 \\
      $w_{RV}$ & 0.2 & $w_{LA}$ & 0.5 & $w_{RA}$ & 0.5 \\
      \bottomrule  % booktabs底部粗线，替代原hline
    \end{tabular}
  \end{threeparttable}
  \vspace{-15pt}  % 与参考格式一致的表末间距
\end{table}

\subsection{Cardiac Standard View Acquisition Experiment}

\label{sec: exp_RL}

\subsubsection{Experimental Setup}

To form the dataset for anatomical feature prior construction, the MM-WHS 2017 dataset \cite{zhuang2019evaluation} was utilized as a static 3D reference. Twelve high-quality cardiac volumetric segmentation samples were selected and projected into A4C images according to ASE guidelines \cite{mitchell2019guidelines}.
These samples were then merged with a clinical A4C segmentation dataset \cite{yang2023graphecho}.
Then, the Gaussian distribution parameters of those anatomical features were fitted using this merged dataset.

The RL training simulation was constructed using 12 high-quality cardiac 3D models from the MM-WHS 2017 dataset \cite{zhuang2019evaluation}.
For each training episode, a cardiac model was randomly selected, and the probe was initialized at the cardiac apex position, paired with a random orientation that enables visualization of an anatomically plausible cardiac section.
% The agent aims to maximize the cumulative reward by adjusting probe posture, ultimately outputting the image that best conforms to the A4C standard \cite{mitchell2019guidelines}.
The double deep Q-network \cite{van2016deep} was adopted for RL training, with parameters in the MDP simulation setup shown in Tab. \ref{Tab: para}.
After training convergence, the RL agent achieved a test success rate of $92.5\%$ with an average of $25.6$ steps in the simulation.

The trained RL agent was directly deployed to the A4C view acquisition experiment, realizing a zero-shot evaluation with no additional fine-tuning on the experimental setup.
This zero-shot deployment is made feasible primarily by the robustness of the anatomical feature prior distributions, allowing them to generalize across sim-to-real discrepancies. 
Meanwhile, the SRG-augmented segmentation model provides anatomically consistent masks of the US image, which ensures the accurate extraction of anatomical features for the RL state and reward calculation in real-time. 
% The experimental platform included a robotic arm (EC66, Cobot Robots), a US imaging device with a probe (MX7, Mindray), an F/T sensor (GLH92003ABO, NBIT), and a cardiac phantom (BPH700-C, Blue Phantom), as shown in Fig. \ref{fig: framework}.
The experimental platform included a robotic arm, a US imaging device with a probe, an F/T sensor, and a cardiac phantom, as shown in Fig. \ref{fig: framework}.
% The system was implemented in C++ on an Intel Core i7-12700 PC with an RTX 4060 GPU.

To verify the practical efficacy of the proposed framework, 15 experiments were performed on the A4C standard view acquisition task under a $7 Hz  $ closed-loop system with random initial probe postures. All experiments were conducted with fixed seeds, unified parameters, and the same trained model to ensure the reliability of results.
The difference between the simulation and practical environment is not only in the imaging cardiac model but also in the initial probe placement constraint. The probe was fixed at the cardiac apex position during simulated trials, while the practical experiments set the probe's initial position to be randomized within the effective imaging range of the A4C view, leading to more realistic variations in cardiac structure visibility. 
Randomized initial probe posture setups were categorized into three deviation levels based on cardiac structure completeness and global polar angle deviation of the initial US image. Mild deviation refers to scenarios where all target structures are visible and $|\phi_\text{all}| \leq 0.5$, moderate deviation denotes cases where either partial structures are missing or $|\phi_\text{all}| > 0.5$, and severe deviation means situations where only one or two target structures are visible and $|\phi_\text{all}| > 0.5$. 
A successful experiment was defined as acquiring an A4C view compliant with ASE guidelines \cite{mitchell2019guidelines}, where all target structures are visible and $|\phi_\text{all}| \leq 0.2$.

\subsubsection{Experimental Results}

\begin{figure}[t]
 \centering
 \vspace{5pt}
 \includegraphics[width=1\linewidth]{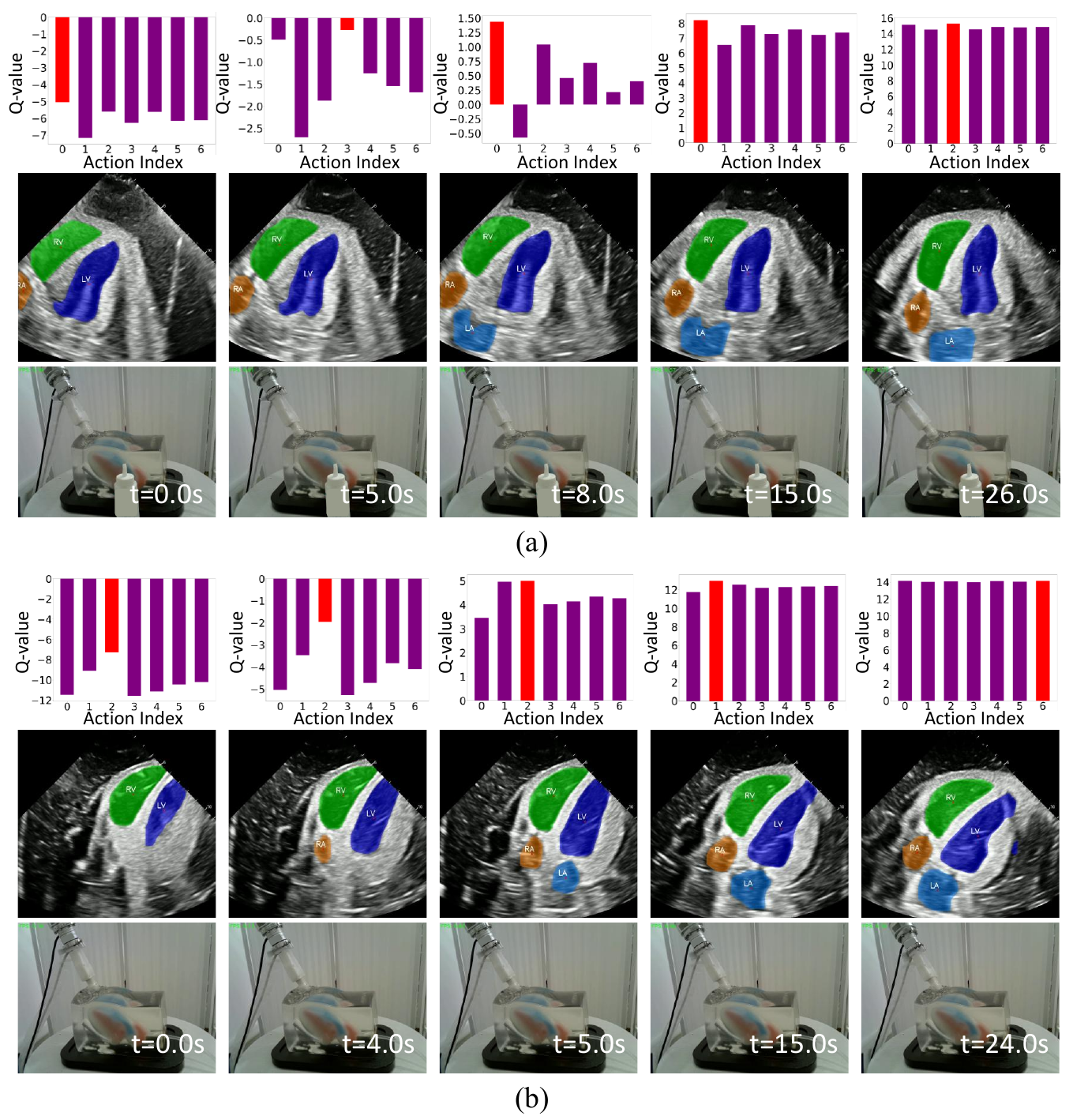}
 \vspace{-20pt}
 \caption{Visualization of two representative experimental results. (a) represents a moderate deviation example, and (b) represents a severe deviation example. In both examples, the content is organized in three rows. The top row illustrates the action Q-value of different action indices, where the red bar represents the selected action. The middle row presents segmented cardiac images, with annotations of cardiac chambers, RV (green), LV (blue), RA (orange), and LA (light blue), overlaid on them. The bottom row displays the corresponding snapshot utilized in the experiment.}
\label{fig: globalScan}
\vspace{-15pt}
\end{figure}

Among the 15 experiments, the model achieved an overall success rate of $86.7\%$ with an average acquisition time of $25.5s$.
For mild deviations, it achieved a success rate of $100\%$ (2 out of 2) with an average acquisition time of 18.5 seconds.
For moderate deviations, the success rate was $85.7\%$ (6 out of 7) with an average time of $31.0s$.
For severe deviations, it achieved a success rate of $83.3\%$ (5 out of 6) with an average time of $21.6s$.
Fig. \ref{fig: globalScan} presents two representative experimental results to validate the model’s practical performance: a moderate deviation setup in (a) and a severe deviation setup in (b).
For the moderate deviation setup, the initial probe posture resulted in a US image where the LA was occluded by US artifacts.
During the probe tuning process, the graph-based segmentation model effectively distinguished the LA from surrounding artifacts.
For the severe deviation setup, the initial US frame only provided incomplete structural information. Despite this, the RL agent evaluated the Q-value of each candidate action and executed a fine-tuning action sequence to converge to an ASE-compliant A4C view in $24.0s$. 
Notably, the system maintained stable performance even when the random position setups led to variations in cardiac orientation and chamber distribution, such as different cardiac orientation tilts and a superiorly shifted RA in Fig. \ref{fig: globalScan}.
The adaptability to handle different conditions stems from the integration of AP modules in not only the semantic segmentation of US images but also the probe adjustment of robotic US scanning, suggesting the ability to handle real-world uncertainties in the proposed framework's autonomous robotic US scanning.

\section{CONCLUSIONS}

This study proposes an AP-driven framework for autonomous robotic cardiac US standard view acquisition.
By injecting APs to unify segmentation and probe adjustment, the framework achieves a closed-loop standard A4C view acquisition. 
The SRG-augmented YOLO segmentation model is proposed to enhance robustness against missed detection, mislabeling between classes, and duplicate prediction issues.
% On the \textit{Special Case} dataset, the SRG-YOLOv11s outperforms its baseline by $11.3\%$ in mAP\textsubscript{50} and $6.8\%$ in mIoU, validating that AP-embedded spatial-topological constraints effectively improve anatomical consistency.
The AP-guided RL probe adjustment uses anatomical feature priors as interpretable benchmarks and reward criteria, achieving zero-shot phantom deployment.
% , achieves a $92.5\%$ success rate in simulation and an $86.7\%$ success rate in phantom experiments with zero-shot phantom deployment, adapting stably to mild ($100\%$ success), moderate ($85.7\%$ success), and severe ($83.3\%$ success) initial probe deviations.
% Such consistent performance across simulated and physical environments demonstrates its ability to bridge the sim-to-real gap through AP-based generalization.

Although the proposed framework achieved promising results, several threads remain for future improvement. The framework has not considered the soft tissue compliance of the skin, which requires the integration of force control mechanisms to improve the reliability of cardiac imaging \cite{zakeri2024ai}. The system has not yet considered cardiac pulsation. Subsequent research will focus on addressing temporal deformation induced by cardiac pulsation and developing the corresponding probe strategies.

\bibliographystyle{IEEEtran}
\bibliography{ref}

\end{document}